\documentclass[runningheads]{llncs}
\usepackage[T1]{fontenc}
\usepackage{graphicx}
\usepackage{cite}
\usepackage{amsmath,amssymb,amsfonts,amsthm}
\usepackage{algorithmic}
\usepackage{textcomp}
\usepackage{xcolor}
\usepackage{multirow}
\usepackage{mathtools}
\usepackage{bm}
\usepackage[ruled,linesnumbered]{algorithm2e}
\usepackage{hyperref}
\usepackage{makecell}
\usepackage{bbding}
\usepackage{tcolorbox}
\usepackage{dsfont}
\usepackage{caption}
\usepackage{subcaption}
\usepackage{xfrac}
\usepackage{wrapfig}

\makeatletter
\newcommand{\thickhline}{%
    \noalign {\ifnum 0=`}\fi \hrule height 1pt
    \futurelet \reserved@a \@xhline
}
\makeatother

\newtcolorbox{rqbox}[3][]
{
  colframe = rq,
  colback  = rqBack,
  coltitle = rq!10,  
  title    = {#3},
  fontupper=\itshape,
  #1,
}

\definecolor{rq}{HTML}{1B365C}
\definecolor{rqBack}{HTML}{9ECBF7}

\def\BibTeX{{\rm B\kern-.05em{\sc i\kern-.025em b}\kern-.08em
    T\kern-.1667em\lower.7ex\hbox{E}\kern-.125emX}}

\newcommand{\tool}{\textsc{FedC$^2$SL}}
\newcommand{\fedpc}{\textsc{FedPC}}
\newcommand{\fedfci}{\textsc{FedFCI}}

\newcommand{\algcomment}[1]{\textcolor{blue}{#1}}

\newcommand{\sm}{Supplementary Material}
\newcommand{\F}{Fig.}
\newcommand{\T}{Table}
\renewcommand{\S}{Sec.}
\newcommand{\A}{Alg.}

\newcommand{\D}{Def.}
\newcommand{\Thm}{Theorem}

\newcommand{\parh}[1]{\smallskip\smallskip\noindent\textbf{#1}}

\newcommand{\ci}{\perp\!\!\!\perp}
\newcommand{\dsep}{\perp\!\!\!\perp_G}

\newcommand{\CBrush}{\textcolor[RGB]{84,130,53}{\Checkmark}}
\newcommand{\XBrush}{\textcolor[RGB]{176,35,24}{\XSolidBrush}}

\begin{document}
\title{Towards Practical Federated Causal Structure Learning}


\author{Zhaoyu Wang\orcidID{0009-0009-6892-1264} \and
Pingchuan Ma$^\dagger$\orcidID{0000-0001-7680-2817} \and
Shuai Wang\orcidID{0000-0002-0866-0308}}
\institute{The Hong Kong University of Science and Technology, Hong Kong SAR\\
\email{1950574@tongji.edu.cn}\\
\email{\{pmaab,shuaiw\}@cse.ust.hk}\\
$^\dagger$corresponding author}

\maketitle              
\begin{abstract}

Understanding causal relations is vital in scientific discovery. The process of
causal structure learning involves identifying causal graphs from observational
data to understand such relations. Usually, a central server performs this task,
but sharing data with the server poses privacy risks. Federated learning can
solve this problem, but existing solutions for federated causal structure
learning make unrealistic assumptions about data and lack convergence
guarantees. \tool\ is a federated constraint-based causal structure learning
scheme that learns causal graphs using a federated conditional independence
test, which examines conditional independence between two variables under a
condition set without collecting raw data from clients. \tool\ requires weaker
and more realistic assumptions about data and offers stronger resistance to data
variability among clients. \fedpc\ and \fedfci\ are the two variants of \tool\
for causal structure learning in causal sufficiency and causal insufficiency,
respectively. The study evaluates \tool\ using both synthetic datasets and
real-world data against existing solutions and finds it demonstrates encouraging
performance and strong resilience to data heterogeneity among clients.

\end{abstract}

\section{Introduction}
\label{ref:intro}

Learning causal relations from data is a fundamental problem in causal
inference. Causal structure learning is a popular approach to identifying causal
relationships in multivariate datasets, represented as a causal graph. This
technique has been successfully applied in various fields such as
medicine~\cite{pinna2010knockouts,shen2020challenges,belyaeva2021dci},
economics~\cite{addo2021exploring}, earth science~\cite{runge2019inferring} and
data analytics~\cite{ma2022xinsight}.

Causal structure learning is performed on a central server with plaintext
datasets. However, in applications like clinical data analysis, data may be
distributed across different parties and may not be shared with a central
server. To address this problem, federated learning is an emerging paradigm that
allows data owners to collaboratively learn a model without sharing their data
in plaintext~\cite{bonawitz2017practical,fereidooni2021safelearn}. However,
current federated learning solutions are designed primarily for machine learning
tasks that aggregate models trained on local datasets.

Several solutions have been proposed for federated causal structure learning
\cite{ng2022towards,gao2021federated,mian2022regret,samet2009privacy}. However,
these solutions have prerequisites that may hinder their general applicability.
For instance, NOTEARS-ADMM \cite{ng2022towards}, which is the state-of-the-art
solution for federated causal structure learning, collects parameterized causal
graphs from clients and uses an ADMM procedure to find the consensus causal
graph in each iteration. Since local graphs jointly participate in the ADMM
procedure, it is non-trivial to employ secure aggregation to protect individual
causal graphs, resulting in a considerable sensitive information leak to the
central server. Additionally, the assumption that data is generated in a known
functional form is deemed unrealistic in many real-life applications. 

In general, many solutions attempt to locally learn a causal graph and aggregate
them together, but this practice is not optimal for federated causal structure
learning. Causal structure learning is known to be error-prone in small
datasets, and local datasets may suffer selection bias with respect to the
global dataset due to the potential heterogeneity of different clients. The
causal graph independently learned from each local dataset may manifest certain
biases with respect to the true causal graph of the whole dataset. 

To address this issue, we propose a novel federated causal structure learning
with constraint-based methods. This paradigm interacts data only with a set of
statistical tests on conditional independence and deduces graphical structure
from the test results. The key idea of our solution is to provide a federated
conditional independence test protocol. Each client holds a local dataset and
computes their local statistics, which are then securely aggregated to derive an
unbiased estimation of the global statistics. With the global statistics, we can
check the global conditional independence relations and conduct constraint-based
causal structure learning accordingly. 

We evaluate our solution with synthetic data and a real-world dataset and
observe better results than baseline federated causal structure learning
algorithms, including state-of-the-art methods
NOTEARS-ADMM~\cite{ng2022towards}, RFCD~\cite{mian2022regret}, and four
voting-based algorithms. Our solution also shows strong resiliency to client
heterogeneity while other baseline algorithms encounter notable performance
downgrades in this setting. Furthermore, our solution facilitates causal feature
selection (CFS) and processes real-world data effectively.

In summary, we make the following contributions: (1) we advocate for federated
causal structure learning with constraint-based paradigms; (2) we design a
federated conditional independence test protocol to minimize privacy leakages
and address client heterogeneity; and, (3) we conduct extensive experiments to
assess the performance of our solution on both synthetic and real-world
datasets. We release our implementation, \tool, on
\url{https://github.com/wangzhaoyu07/FedC2SL} for further research and
comparison.
\section{Preliminary}

In this section, we review preliminary knowledge of causal structure learning. 

\noindent\textbf{Notations.} Let $X$ and $\bm{X}$ represent a variable and a set
of variables, respectively. In a graph, a node and a variable share the same
notation. The sets of nodes and edges in a causal graph $G$ are denoted as $V_G$
and $E_G$, respectively. The notation $X \to Y\in E_G$ indicates that $X$ is a
parent of $Y$, while $X \leftrightarrow Y\in E_G$ indicates that $X$ and $Y$ are
connected by a bidirected edge. The sets of neighbors and parents of $X$ in $G$
are denoted as $N_G(X)$ and $Pa_G(X)$, respectively. The notation
$[K]\coloneqq\{1,\cdots,K\}$ is defined as the set of integers from 1 to $K$.

\subsection{Causal Structure Learning}
\label{subsec:csl}

In causal inference, the relationship between data is often presented as a
causal graph, which can take the form of a directed acyclic graph (DAG) or
maximal ancestral graph (MAG). These two representations are used to depict
causal relationships under different assumptions. In the following paragraphs,
we introduce the corresponding causal graphs and formalize these canonical
assumptions.

\begin{wrapfigure}{r}{0.5\textwidth}
  \vspace{-10pt}
  \centering
\includegraphics[width=0.5\columnwidth]{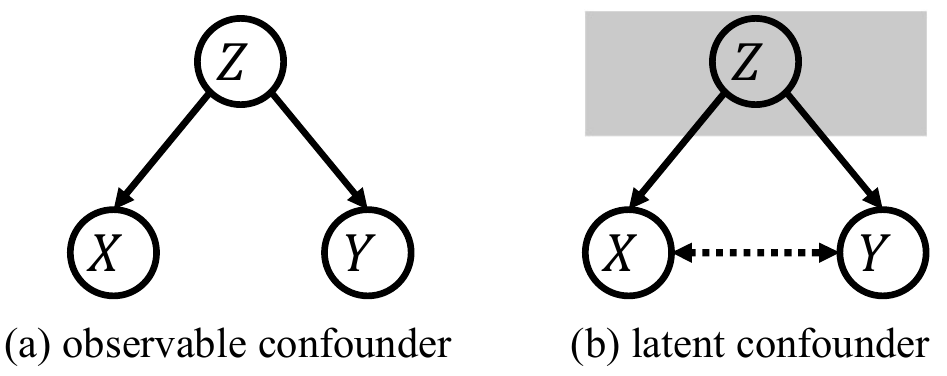}
\vspace{-10pt}
\caption{Examples of observable and latent confounders.}
\label{fig:causal-graph}
\vspace{-15pt}
\end{wrapfigure}

\parh{Graphical Representation.}~Causal relations among variables in a
multivariate dataset can be depicted using a causal graph. The causal graph can
either be a directed acyclic graph (DAG), where adjacent variables are connected
by a directed edge, or a mixed acyclic graph (MAG), which allows for bidirected
edges to indicate shared latent confounders between two variables. In the DAG
format, if a latent confounder is not observed, statistical associations between
variables can exist without their true causal relations being well-represented.
MAG overcomes this shortcoming and can be constructed from a true DAG and a set
of latent variables, using a set of construction criteria
\cite{zhang2008completeness}. See \F~\ref{fig:causal-graph}~(a) for an example
of a DAG depicting a directed edge ($Z\to X$). 

\parh{Causal Sufficiency.}~Learning a DAG assumes a causally sufficient set of
variables~\cite{spirtes2000causation}. $\bm{X}$ is causally sufficient if there
is no hidden cause $Z\notin\bm{X}$ causing more than one variable in $\bm{X}$.
However, real-world data may not satisfy this assumption. MAG addresses this
issue by introducing a bidirected edge $\leftrightarrow$. See
\F~\ref{fig:causal-graph}~(b) for an example where a bidirected edge between
$X,Y$ due to the absence of $Z$.

\parh{Global Markov Property (GMP).}~The Global Markov Property
(GMP)~\cite{lauritzen1996graphical} connects graphical structures and
statistical properties. It can be stated as: $X\dsep Y\mid \bm{Z} \implies X\ci
Y\mid \bm{Z}$.
Here, $\dsep$ represents graphical separation and $\ci$ represents statistical
conditional independence in the joint probability distribution $P_{\bm{X}}$.
D-separation is a structural constraint for directed acyclic graphs (DAGs),
while m-separation is a constraint for mixed graphs (MAGs). We present their
definitions in \sm.

\parh{Faithfulness Assumption.}~Faithfulness assumption states that conditional
independence on the joint distribution implies d-separation (or m-separation) on
the causal graph. Formally, $X\ci Y\mid \bm{Z}\implies X\dsep Y\mid \bm{Z}$


In the remainder of the paper, we assume GMP and faithfulness assumption always
hold. Moreover, we assume causal sufficiency in \fedpc\ and propose \fedfci\
that is also tolerant to causally insufficient data.

\parh{Markov Equivalence Class (MEC).}~Given the Markov condition and
faithfulness assumption, statistical tests can be performed on data to deduce
graph structures through graphical separation constraints. However, inferring
the full structure of a causal graph from data is difficult and can lead to
multiple causal graphs being compatible with the constraints deduced from
conditional independence. To address this, causal structure learning algorithms
aim to recover a MEC, which summarizes a set of causal graphs sharing the same
set of d-separations (or m-separations)~\cite{peters2017elements}. The MEC is
represented as a CPDAG for DAG learning and as a PAG for MAG
learning~\cite{zhang2008completeness}. \tool\ follows standard
conventions~\cite{peters2017elements,zhang2008completeness,ma2022ml4s} in
recovering the MEC of a given dataset.

\parh{Constraint-based Causal Structure Learning.}~Constraint-based methods are
commonly used for causal structure learning, identifying the MEC from
observational datasets. The PC algorithm (see details in \sm) is a representative
constraint-based causal structure learning algorithm. 
This algorithm involves two phases: learning the causal skeleton and orienting 
the edges. During the former phase, the adjacency relations between variables 
are learned and an undirected graph is created. In this graph, the edges represent 
the underlying causal graph's skeleton. In the latter phase, a set of orientation 
rules is applied to assign a causal direction to the undirected edges of the skeleton.
In comparison to the PC algorithm, which performs DAG learning, the FCI algorithm~\cite{zhang2008completeness} 
is designed for MAG learning, incorporating another set of orientation rules while using 
a similar skeleton learning procedure of the PC algorithm.

\section{Research Overview}
\label{sec:overview}

This section presents the research overview. We begin by providing an overview
of \tool\ in \S~\ref{subsec:setup}, covering the problem setup and threat model.
\S~\ref{subsec:cmp} provides a comparison between our solution and existing
approaches.

\subsection{Problem Setup}
\label{subsec:setup}

In this paper, we consider two causal discovery problems: \fedpc\ and \fedfci.

\parh{\fedpc.}~Assuming causal sufficiency, \fedpc\ involves a DAG $G=(V,E)$
that encodes the causal relationships among a variable vector
$\bm{X}=\{X_1,\cdots,X_d\}$ with a joint probability distribution $P_{\bm{X}}$
satisfying the Global Markov Property (GMP) with respect to $G$, and $G$ is
faithful to $P_{\bm{X}}$.

\parh{\fedfci.}~In causal insufficient data, \fedfci\ involves a MAG
$M=(V,E)$ that encodes the causal relationships among a variable vector
$\bm{X}=\{X_1,\cdots,X_d\}$ with a joint probability distribution $P_{\bm{X}\cup
L}$, where $L$ is a set of unknown latent variables. Here, $P_{\bm{X}}$ is the
observable distribution with $P_{\bm{X}\cup L}$ being marginalized on $L$. If
$L$ is empty, then the setting is equivalent to \fedpc. We assume that
$P_{\bm{X}}$ satisfies GMP with respect to $G$ and $M$ is faithful to
$P_{\bm{X}}$.

We now describe the setting of clients that are identical for either \fedpc\ or
\fedfci. Suppose that there are $K$ local datasets
$\mathcal{D}\coloneqq\{D^1,\cdots,D^K\}$ and
$D^i=\{\bm{x}^i_1,\cdots,\bm{x}^i_{n_i}\}$. We denote $\bm{x}^i_{j,k}$ be the
$k$-th element of the $j$-th record in the $i$-th local dataset. Each record in
the global dataset $\mathcal{D}$ is sampled \textit{i.i.d.} (independent and
identically distributed) from $P_{\bm{X}}$. We allow for selection bias on
client local datasets as long as the global dataset is unbiased with respect to
$P_{\bm{X}}$, which is one of the main challenges in federated
learning~\cite{kairouz2021advances}. For example, local datasets from different
hospitals may be biased on different patient subpopulations. However, with a
sufficient number of clients, the global dataset (by pooling all local datasets)
becomes unbiased. This assumption is weaker than the Invariant DAG Assumption in
DS-FCD~\cite{gao2021federated}. We borrow the concept from general causal
structure learning~\cite{versteeg2022local} and formally define this property as
follows.

\begin{definition}[Client Heterogeneity]
  \label{def:hetro}
  Let $\{X_1,\cdots,X_d\}$ be the visible variables in the dataset and $G$ be
  the ground-truth causal graph. To represent client-wise heterogeneity, we
  assume that there is an implicit surrogate variable $C: [K]$ be the child
  variable of $S\subseteq \{X_1,\cdots,X_d\}$ in an augmented causal graph and
  the i-th client holds the records with $C=i$. When $S=\emptyset$, the local
  datasets are homogeneous.
\end{definition}

The overall goal is to recover MEC of $G$ or $M$ from distributed local datasets
in the presence of client heterogeneity while minimizing data leakages.

\parh{Threat Model.}~Our threat model aligns with the standard
setting~\cite{yang2019federated}. We assume that all parties including the
central server and clients are honest but curious, meaning that they will follow
the protocol but are interested in learning as much private information as
possible from others. We are concerned with the leakage of private client data,
and we do not consider any coalitions between participants. We will show later
that \tool\ is resilient to client dropouts, although we do not explicitly
consider this during algorithm design.

\parh{Security Objective.}~The federated learning paradigms aim to prevent raw
data sharing, and only aggregated results are released \cite{wang2021fed}. We
aim to achieve MPC-style security to ensure that the semi-honest server only
knows the aggregated results and not individual updates. To establish this
property formally, we define client indistinguishability in federated causal
structure learning. 

\begin{definition}[Client Indistinguishability]
    Let $\bm{x}\in D^i$ be a record that only exists in the $i$-th client (i.e.,
    $\forall j\neq i,\forall \bm{x}'\in D^j, \bm{x}\neq \bm{x}'$). Let
    $\mathbb{P}(A)$ be the public knowledge (e.g., intermediate data and final
    results) revealed in the protocol $A$. $A$ is said to be \textit{client
    indistinguishable} for an adversary if\, $\forall i,j\in [K], P(\bm{x}\in
    D^i\mid \mathbb{P}(A))=P(\bm{x}\in D^j\mid \mathbb{P}(A))$.
\end{definition}



\subsection{Comparison with Existing Solutions}
\label{subsec:cmp}

In this section, we review existing solutions and compare them with \tool. We
summarize existing works and \tool, in terms of assumptions, application scope,
and leakage, in \T~\ref{tab:comparison}.

\begin{table*}[!ht]
  \vspace{-15pt}
  \centering
  \caption{Comparing existing works and \tool.} 
  \label{tab:comparison}
\resizebox{\linewidth}{!}{
\begin{tabular}{l|c|c|c|c|c|c}
    \hline
    \textbf{Solution} & \textbf{Input}  & \textbf{Assumption} & \textbf{Heterogeneity} & \textbf{Application Scope}  & \textbf{Client Leakage} & \textbf{Global Leakage}\\ 
  \hline
  NOTEARS-ADMM~\cite{ng2022towards} & Data & Additive Noise & \XBrush & DAG& Individual Graph + Parameter  & Graph + Parameter \\
  \hline
  DS-FCD~\cite{gao2021federated} & Data & Additive Noise & \XBrush & DAG& Individual Graph + Parameter & Graph + Parameter \\
  \hline
  RFCD~\cite{mian2022regret}& Data  & Additive Noise & \XBrush & DAG& Individual Graph Fitness & Graph \\
  \hline
  K2~\cite{samet2009privacy}& Data + Order & Faithfulness & \CBrush & DAG& Aggregated Fitness & Graph  \\
  \hline
  \tool\ & Data    & Faithfulness & \CBrush & DAG \& MAG& Aggregated Low-dim. Distribution  & Graph \\
  \hline
  \end{tabular}
}
\vspace{-15pt}
\end{table*}

\parh{Comparison of Prerequisites.} Most federated causal structure learning
solutions assume additive noise in the data generating process, which is
considered stronger than the faithfulness assumption in
K2~\cite{samet2009privacy} and \tool. However, K2 requires prior knowledge of
the topological order of nodes in a ground-truth DAG, which is impractical.
Additionally, solutions that learn local graphs independently are intolerant of
client heterogeneity, as their performance would degrade arbitrarily in theory.
\tool\ is the only solution that supports MAG learning, allowing for causal
structure learning on causally insufficient data, making it a more practical
option. (See \S~\ref{subsec:csl} for more details.)

\parh{Comparison on Privacy Protection.}~We review the privacy protection
mechanisms in proposed solutions. On the client side,
NOTEARS-ADMM~\cite{ng2022towards} and DS-FCD~\cite{gao2021federated} require
clients to update the local causal graph and corresponding parameters in
plaintext to the global server. These graph parameters consist of multiple
regression models trained on local datasets, which are vulnerable to privacy
attacks on ML models. RFCD~\cite{mian2022regret} requires clients to send the
fitness score of global causal graphs on the local dataset, which poses a
privacy risk for adversaries to infer the source of particular data samples and
violate the client indistinguishability. In contrast, K2~\cite{samet2009privacy}
and \tool\ use secure aggregation or secure multi-party computation protocols
such that only aggregated results are revealed. K2 employs a score function to
measure the fitness of a local structure and the global structure is established
by selecting the best local structure in a greedy manner. The score function is
computed over the distributed clients with MPC schemes such that individual
updates are protected. \tool\ uses a constraint-based strategy to learn the
causal graph and securely aggregates the data distribution marginalized over
multiple low-dimensional subspaces to the global server. The marginalized
low-dimensional distributions are strictly less informative than NOTEARS-ADMM
and DS-FCD. The global server asserts conditional independence on the aggregated
distributions and deduces graphical separations by faithfulness accordingly.

\parh{Asymptotic Convergence.}~\tool\ is inherited from constraint-based methods
that offers asymptotic convergence to the MEC of the ground-truth causal graph
under certain assumptions. In contrast, NOTEARS-ADMM and DS-FCD use continuous
optimization to the non-convex function and only converge on stationary
solutions. RFCD and K2 use greedy search over the combinatorial graph space,
which does not provide global convergence guarantees.

\section{\tool}

In this section, we present \tool, a novel federated causal structure learning
algorithm with minimized privacy leakage compared to its counterparts.

\subsection{Causal Structure Learning}

As discussed in \S~\ref{subsec:csl}, the causal graph is learned by testing
conditional independence in the dataset. A MEC of the causal graph contains all
conditional independence, as per GMP and faithfulness assumption. Moreover, the
MEC can be recovered from the set of all conditional independence relations.
Therefore, the set of all conditional independence relations in a dataset is
both necessary and sufficient to represent the MEC of the underlying causal
graph.

\begin{remark}
  Under GMP and faithfulness assumption, a Markov equivalence class of causal
  graph encodes all conditional independence relations among data. Once the
  Markov equivalence class is revealed, all conditional independence relations
  are revealed simultaneously. Therefore, the conditional independence stands
  for the minimal information leak of federated causal structure learning.
\end{remark}

Instead of creating specific federated causal structure learning methods, we
propose using a federated conditional independence test procedure. This
procedure is fundamental to all constraint-based causal structure learning
algorithms, such as the PC algorithm. By implementing our federated version, we
can replace the centralized conditional independence tester in any existing
constraint-based causal structure learning algorithm and make it federated. In
this paper, we apply our federated conditional independence test procedure to
two well-known causal structure learning algorithms, namely, \fedpc\ and
\fedfci, which are based on the PC algorithm~\cite{spirtes2000causation} and FCI
algorithm~\cite{spirtes2000causation,zhang2008completeness}, respectively.

\subsection{Federated Conditional Independence Test}

To enhance privacy protection, a multiparty secure conditional independence test
that releases only the conditional independence relations would be ideal.
However, implementing such a solution using MPC would result in impractical
computational overheads for producing real-world datasets. Therefore, we propose
a practical trade-off that boosts computation efficiency while causing
negligible privacy leakage on relatively insensitive information.

To introduce our federated conditional independence test protocol, we first
explain how to test conditional independence in a centralized dataset. Consider
three random variables $X$, $Y$, and $Z$ from a multivariate discrete
distributions. The conditional independence of $X$ and $Y$ given $Z$ is defined
as follows:

\begin{definition}[Conditional Independence]
  $X$ and $Y$ are conditionally independent given $Z$ if and only if, for all
  possible $(x,y,z)\in (X,Y,Z)$, $P(X=x,Y=y|Z=z) = P(X=x|Z=z)P(Y=y|Z=z)$.
  \label{def:ci}
\end{definition}

While \D~\ref{def:ci} is straightforward to verify, it is non-trivial to
statistically test this property with finite samples. The most popular way is to
use $\chi^2$-test~\cite{bishop1976discrete}, whose null hypothesis and
alternative hypothesis are defined as follows:

\begin{equation}
  H_0: X\ci Y|Z, H_1: X\not\ci Y|Z
\end{equation}

The statistic $\hat{Q}$ is computed as $\hat{Q} = \sum_{x,y,z}
\frac{(v_{xyz}-\frac{v_{xz}v_{yz}}{v_z})^2}{\frac{v_{xz}v_{yz}}{v_z}}$ where
$v_{xyz}$ is the number of samples with $X=x$, $Y=y$ and $Z=z$; and so on. Under
null hypothesis $H_0$, $\hat{Q}$ follows a $\chi^2_{\texttt{dof}}$ distribution
where $\texttt{dof}=\sum_{z\in Z}(|X_{\bm{Z}=z}|-1)(|Y_{\bm{Z}=z}|-1)$ is the
degree of freedom and $|X|,|Y|$ denote the cardinality of the multivariate
discrete random variable. Let $1-\alpha$ be the significance level. The null
hypothesis is rejected if $\hat{Q}>\chi^2_{\texttt{dof}}(1-\alpha)$. 


\parh{Why a Voting Scheme Is Not Suitable?}~One potential approach to test
conditional independence in a federated setting is to perform standard
$\chi^2$-tests on each client independently and use the voted local conditional
independence as the conditional independence on the global dataset. However,
this approach is not feasible for two reasons. Firstly, the $\chi^2$-test
requires that all $v_{xyz}$ are larger than 5 to ensure its
validity~\cite{triola2014essentials}. This requirement is often unattainable on
small local datasets, leading to inaccurate test results. Secondly, even if the
requirement is met, the voting result may not reflect the global conditional
independence in the presence of selection bias on the client dataset. As will be
shown in \S~\ref{sec:eval}, simple voting strategies often yield inaccurate
results.



To preserve privacy while computing $\hat{Q}$ on the global dataset, we can
perform secure aggregation over the four counts ($v_z,v_{xz},v_{yz},v_{xyz}$)
instead of using the voting scheme. Securely summing up
$v_{xyz}^1,\cdots,v_{xyz}^K$ from all clients can obtain $v_{xyz}$. However, if
$Z$ contains multiple variables, releasing $v_{xyz}$ could raise privacy
concerns due to its encoding of the joint distribution of multiple variables. We
discuss the privacy implications of releasing such high-dimensional distribution
in the following paragraph.

\parh{High-Dim.~Distribution vs.~Low-Dim.~Distribution.}~We note that
high-dimensional distribution is more sensitive than low-dimensional
distribution, which allows adversaries to localize a particular instance (e.g.,
patient of a minority disease). Hence, we anticipate to avoid such leakages. In
contrast, the joint distribution marginalized over low-dimensional subspace is
generally less sensitive. It can be deemed as a high-level summary of data
distribution and individual privacy is retained on a reasonable degree.

\begin{algorithm}[H]
  \footnotesize
  \caption{$\texttt{Fed-CI}(X\ci Y\mid \bm{Z})$}
\label{alg:cit}
\KwIn{Data in $K$ clients: $\mathcal{D}\coloneqq\{D^1,\cdots,D^K\}$;
Statistical Significance: $1-\sigma$.}
\KwOut{Whether reject $X\ci Y\mid \bm{Z}$.}
\lIf{$\bm{Z}=\emptyset$}{$\bm{Z}\leftarrow\{\mathds{1}\}$}
\ForEach{$z\in \bm{Z}$}{
    \algcomment{// i) compute marginal distribution}\\
    \algcomment{// client side:}\\
    let $v^i_z$ be the count of $\bm{Z}=z$ on $D^i$\;
    let $v_{xz}^i,v_{yz}^i$ be the count of $X=x$ (or $Y=y$) with $\bm{Z}=z$ on $D^i$\;
    \algcomment{// server side:}\\
    $v_z\leftarrow \texttt{SecureAgg}(\{v^i\}_{i\in [K]})$\;
    \lForEach{$x\in X$}{$v_x\leftarrow\texttt{SecureAgg}(\{v^i_{xz}\}_{i\in [K]})$}
    \lForEach{$y\in Y$}{$v_y\leftarrow\texttt{SecureAgg}(\{v^i_{yz}\}_{i\in [K]})$}
    \lForEach{$(x,y)\in X,Y$}{broadcast $\bar{v}_{xyz}=\frac{v_{xz}v_{yz}}{v_z}$}
    sample $\bm{P}$ from $\mathcal{Q}^{l\times m}_{2,0}$ and broadcast $\bm{P}$\;
    \algcomment{// ii) compute $\chi^2$ statistics}\\
    \algcomment{// client side:}\\
    $\bm{u}_z^i[\mathbb{I}(x,y)]\leftarrow\frac{v_{xyz}^i-\frac{\bar{v}_{xyz}}{K}}{\sqrt{\bar{v}_{xyz}}}$\;
    $\bm{e}^i\leftarrow \bm{P}\times \bm{u^i_z}$\;
    \algcomment{// server side:}\\
    $\bm{e}\leftarrow\texttt{SecureAgg}(\{\bm{e}^i\}_{i\in [K]})$\;
    $\hat{Q}_z\leftarrow\frac{\sum_{k=1}^l |e_k|^{2/l}}{(\frac{2}{\pi}\Gamma(\frac{2}{l})\Gamma(1-\frac{1}{l})\sin(\frac{\pi}{l}))^l}$\;
    $\texttt{dof}_z\leftarrow (|X_{\bm{Z}=z}|-1)(|Y_{\bm{Z}=z}|-1)$\; 
}
\algcomment{// iii) aggregate $\chi^2$ statistics}\\
$\hat{Q}\leftarrow\sum \hat{Q}_z$\;
$\texttt{dof}\leftarrow\sum \texttt{dof}_z$\;
\lIf{$\hat{Q}>\chi^2_{\texttt{dof},1-\sigma}$}{\Return{reject null hypothesis}}
\lElse{\Return{fail to reject null hypothesis}}
\end{algorithm}

To alleviate the direct release of high-dimensional distributions, we leverage
the idea in~\cite{wang2021fed} to recast $\hat{Q}$ statistic into a second
frequency moment estimation problem and employ random projection to hide the
distribution. Specifically, let $\bar{v}_{xyz}=\frac{v_{xz}v_{yz}}{v_z}$. For
each client, we compute
$\bm{u}_z^i[\mathbb{I}(x,y)]=\frac{v_{xyz}^i-\frac{\bar{v}_{xyz}}{K}}{\sqrt{\bar{v}_{xyz}}}$
where $\mathbb{I}: [|X|]\times[|Y|]\to [|X||Y|]$ is an index function. The
$\hat{Q}$ can be rewritten as

\begin{equation}
  \begin{aligned}
    \hat{Q}&=\sum_{x,y,z} \frac{(v_{xyz}-\frac{v_{xz}v_{yz}}{v_z})^2}{\frac{v_{xz}v_{yz}}{v_z}} =\sum_{z}\sum_{x,y}\left(\frac{v_{xyz}-\bar{v}_{xyz}}{\sqrt{\bar{v}_{xyz}}}\right)^2\\
    &=\sum_{z} \Vert \sum_{i\in [K]}\bm{u}^i_z\Vert_2^2 =\sum_{z} \Vert \bm{u}_z\Vert_2^2
  \end{aligned}
\end{equation} 

It is worth noting that the above recasting does not obviously conceal $v_{xyz}$
because it can still be derived from $\bm{u}_z$. To protect $\bm{u}_z^i$, a
random projection is employed to encode $\bm{u}_z^i$ into $\bm{e}^i$ and a
geometric mean estimation is performed over the encoding. Then, the main result
of~\cite{wang2021fed,li2008estimators} implies the following theorem.

\begin{theorem}
  \label{thm:encoding}
  Let $\bm{P}$ be a projection matrix whose values are independently sampled
  from a $\alpha$-stable distribution~\cite{indyk2006stable}
  $\mathcal{Q}^{l\times m}_{2,0}$ ($m=|X||Y|$), $\bm{e}^i=P\times \bm{u}^i_z$ be
  the encoding on the i-th client and $\bm{e}=\sum_{i\in [K]} \bm{e}^i$ be the
  aggregated encoding. $\hat{d}_{(2),gm}=\frac{\sum_{k=1}^l
  |e_k|^{2/l}}{(\frac{2}{\pi}\Gamma(\frac{2}{l})\Gamma(1-\frac{1}{l})\sin(\frac{\pi}{l}))^l}$
  is the unbiased estimation on $\Vert \bm{u}_z\Vert_2^2$.
\end{theorem}

Accordingly, we can compute $\hat{Q}_z$ for each $z\in Z$ and sum them up to
obtain $\hat{Q}$. Using secure aggregation, the joint distribution of $X,Y,Z$ on
local datasets is already perfectly invisible to the central server. The
encoding scheme in the above theorem provides additional privacy protection to
the distribution on the global dataset. Specifically, under appropriate
parameters, after receiving the aggregated encoding $\bm{e}$, the server cannot
revert back to the original $\bm{u}_z$. Indeed, given $\bm{e}$, $\bm{u}_z$ is
concealed into a subspace with exponential feasible solutions according to
\Thm~2 in~\cite{wang2021fed}. We now outline the workflow of our federated
conditional independence test protocol in \A~\ref{alg:cit}. To make
\A~\ref{alg:cit} compatible to empty condition set (i.e., $\bm{Z}=\emptyset$),
we add a dummy variable $\mathds{1}$ to $\bm{Z}$ (line 1) and the subsequent
loop (lines 2--21) only contains one iteration applied on the entire (local)
datasets (e.g., $v^i_z$ is the count of total samples in $D^i$, and so on). In
each iteration where a possible value of $\bm{Z}$ is picked, each client counts
$v^i_z,v^i_{xz},v^i_{yz}$ privately (lines 5--6) and securely aggregates to the
server (lines 8--10). The server then broadcasts
$\bar{v}_{xyz}=\frac{v_{xz}v_{yz}}{v_z}$ for each $(x,y)\in X,Y$ and the
projection matrix $\bm{P}$ to all clients (lines 11-12). Then, the client
computes $\bm{u}_z^i$ and generates $\bm{e}^i$ (lines 15--16). The server
aggregates encodings (line 18), perform geometric mean estimation to derive
$\hat{Q}_z$ (line 19) and compute degree of freedom $\texttt{dof}_z$ (line 20).
After enumerating all $z\in Z$, the total $\chi^2$ statistics and the total
degree of freedom is computed by summing $\hat{Q}_z$ and $\texttt{dof}_z$ up,
respectively (lines 23--24). Finally, $\hat{Q}$ is compared against
$\chi^2_{\texttt{dof},1-\sigma}$ and \A~\ref{alg:cit} decides whether to reject
null hypothesis (lines 25--26).

\section{Evaluation}
\label{sec:eval}
In this section, we evaluate \tool\ to answer the following three research
questions (RQs): \textbf{RQ1: Effectiveness.}~Does \tool\ effectively recover
causal relations from data with different variable sizes and client numbers?
\textbf{RQ2: Resiliency.}~Does \tool\ manifest resiliency in terms of client
dropouts or client heterogeneity? \textbf{RQ3: Real-world Data.}~Does \tool\
identify reasonable causal relations on real-world data? We answer them in the
following sections.


\subsection{Experimental Setup}


\parh{Baselines.}~We compare the performance of \tool\ with seven baselines,
including two SOTA methods: NOTEARS-ADMM~\cite{ng2022towards} and
RFCD~\cite{mian2022regret}. We also implement two baseline algorithms, PC-Voting
and PC-CIT-Voting, which aggregate and vote on local causal graphs to form a
global causal graph. Additionally, we compare with the centralized PC
algorithm~\cite{spirtes2000causation}, as well as FCI algorithm and two
voting-based baselines (FCI-Voting and FCI-CIT-Voting). We report the
hyperparameters in \sm.

\parh{Dataset.}~We evaluate \tool\ on synthetic and real-world datasets. We
describe the generation of synthetic datasets in \sm. We use the discrete
version of the Sachs dataset~\cite{sachs2005causal}, a real-world dataset on
protein expressions involved in human immune system cells.

\parh{Metrics.}~We use Structural Hamming Distance (SHD) between the Markov
equivalence classes of learned causal graph and the ground truth (lower is
better). We also record the processing time. For each experiment, we repeat ten
times and report the averaged results.

\subsection{Effectiveness}
\label{subsec:rq1}

\begin{figure*}
  \centering
  \begin{subfigure}[b]{0.49\columnwidth}
      \centering
      \includegraphics[width=\columnwidth]{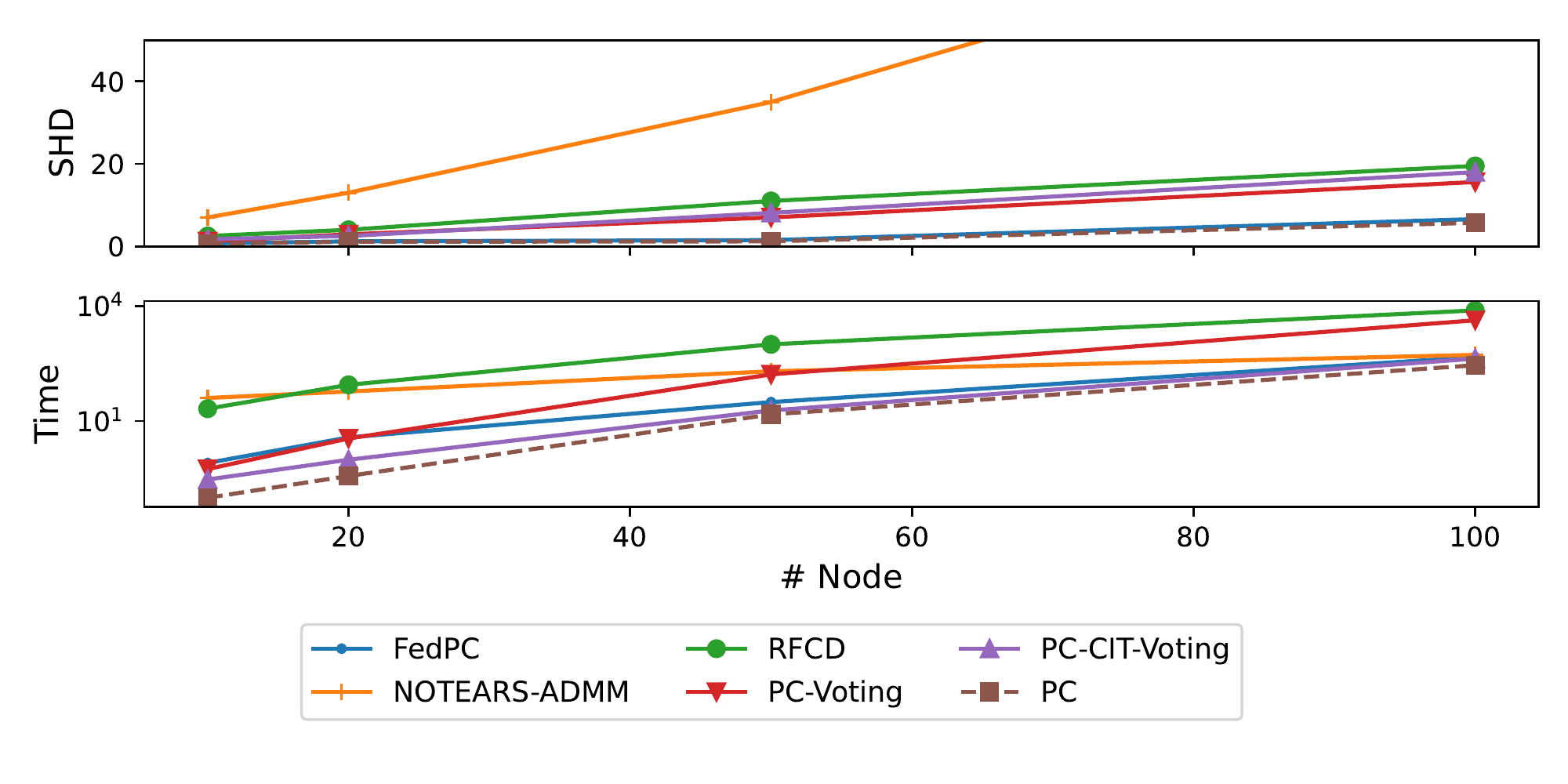}
  \end{subfigure}
  \hfill
  \begin{subfigure}[b]{0.49\columnwidth}
      \centering
      \includegraphics[width=\columnwidth]{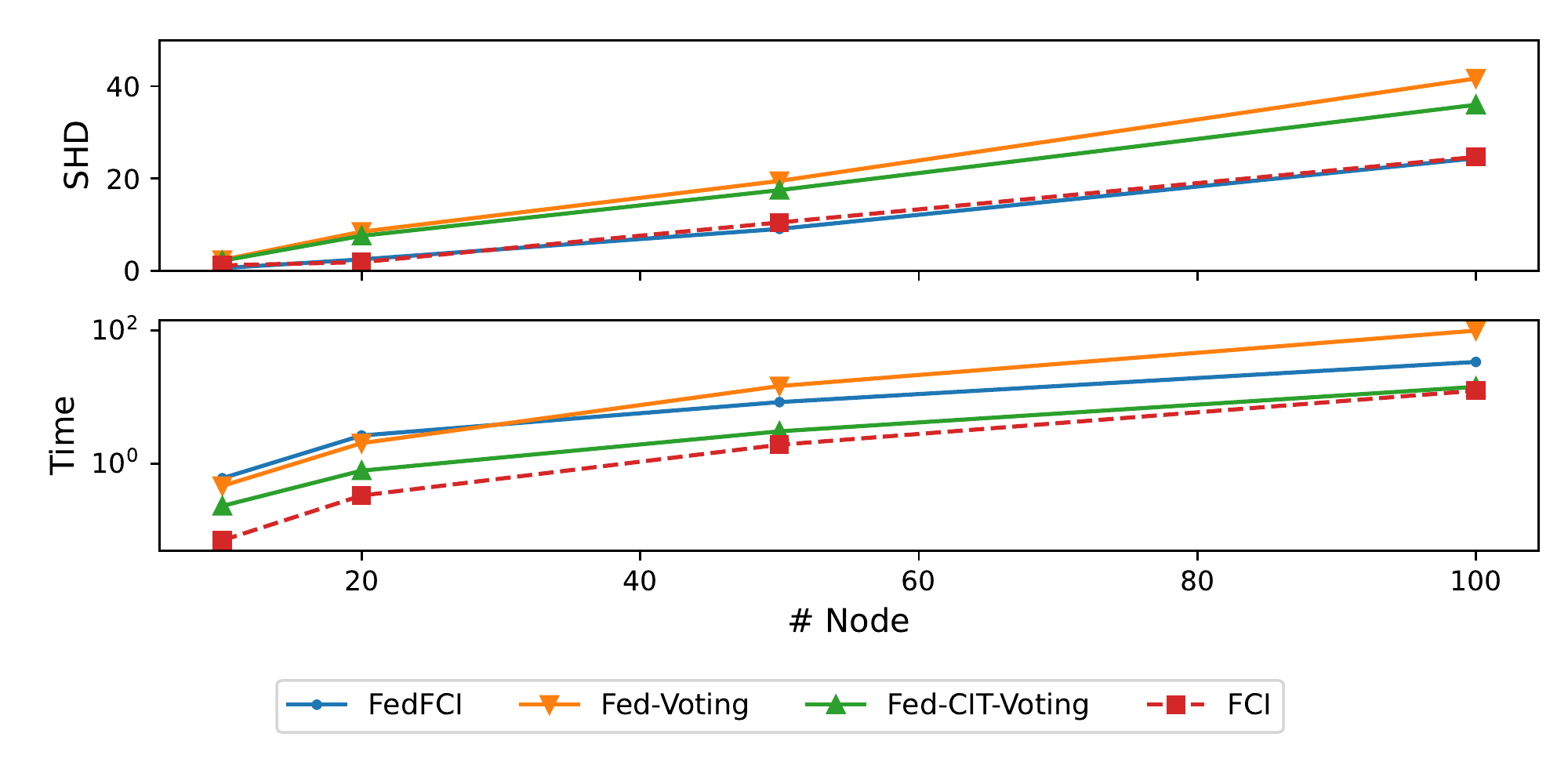}
  \end{subfigure}
  \hfill

  \begin{subfigure}[b]{0.49\columnwidth}
      \centering
      \includegraphics[width=\columnwidth]{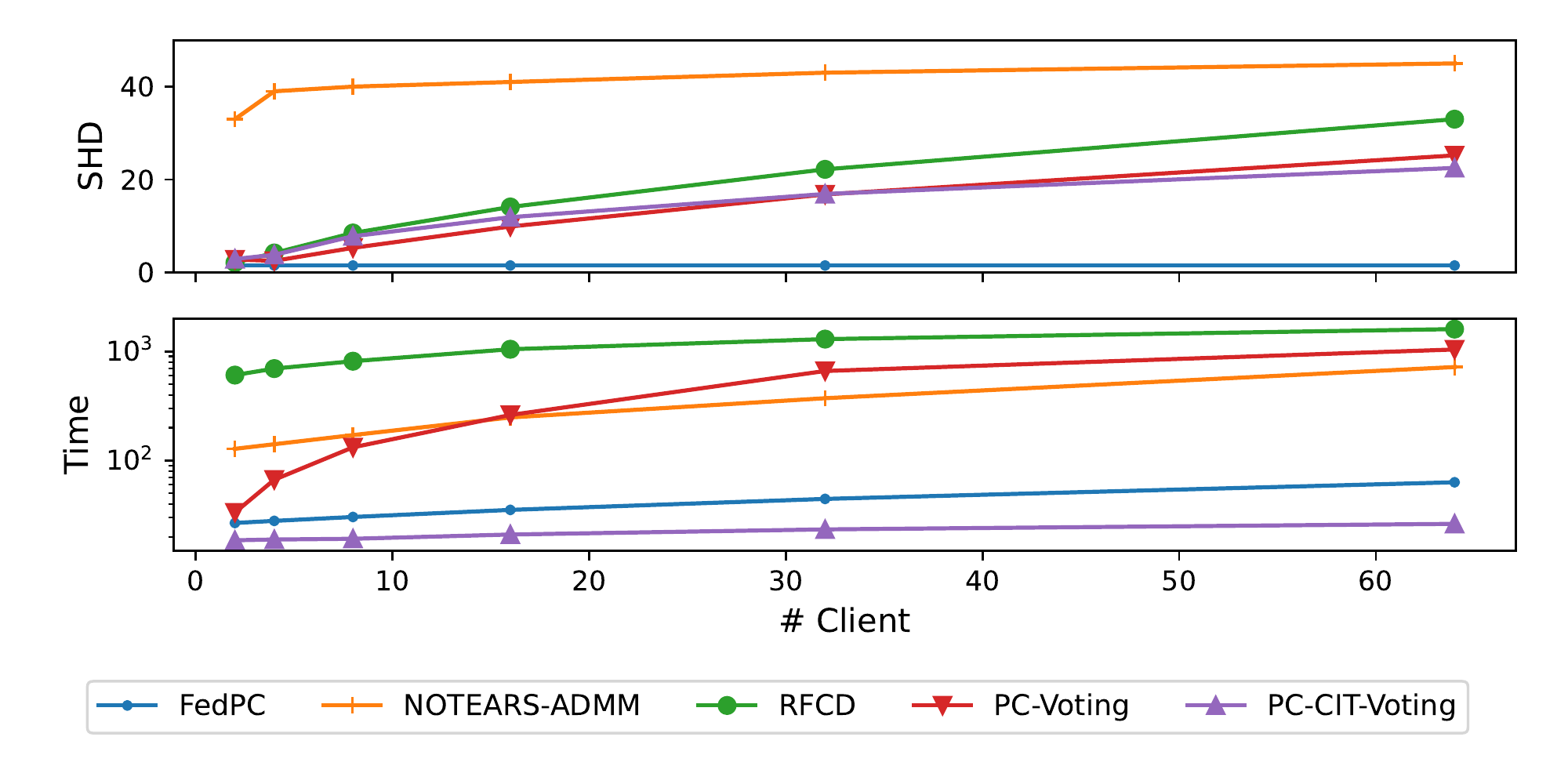}
  \end{subfigure}
  \hfill
  \begin{subfigure}[b]{0.49\columnwidth}
      \centering
      \includegraphics[width=\columnwidth]{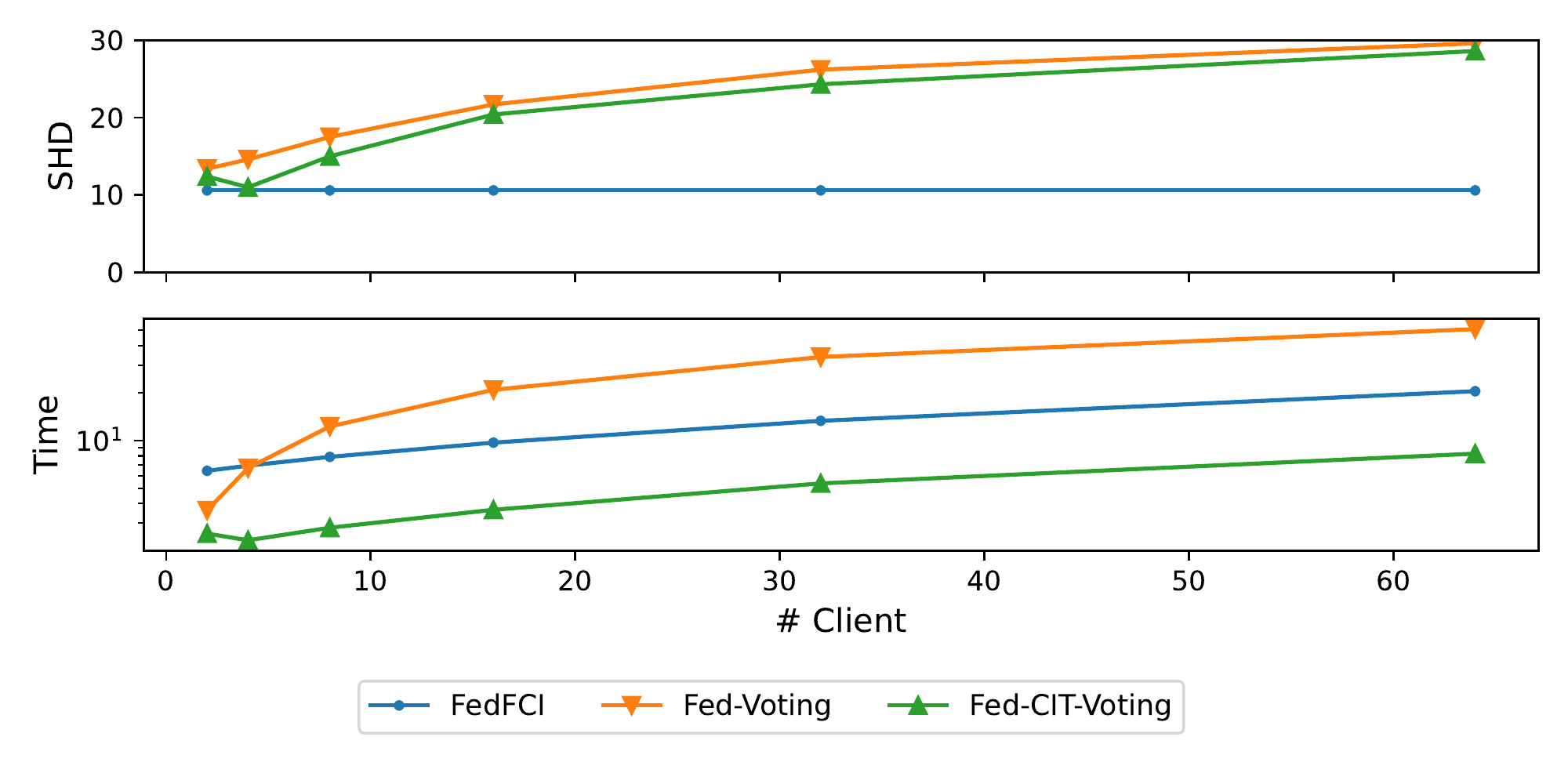}
  \end{subfigure}
     \caption{Performance on different variable sizes and client numbers.}
     \label{fig:rq1}
\end{figure*}

High-dimensional datasets pose challenges for causal structure learning. We
evaluate the performance of \tool\ on datasets with varying variable sizes and
fixed client size ($K=10$) in \F~\ref{fig:rq1}. We report the results for
federated causal structure learning for DAG and MAG. We observe that \tool\
consistently outperforms all other methods (excluding its centralized version)
in terms of SHD on all scales. Its accuracy is closely aligned with PC and FCI
(i.e., its centralized version), indicating negligible utility loss in the
federated procedure.

Furthermore, the processing time of \tool\ is slightly higher but acceptable and
often lower than other counterparts. On datasets with 100 variables, \tool\
shows a much lower SHD than other federated algorithms, indicating its
scalability to high-dimensional data. In contrast, NOTEARS-ADMM performs poorly
on datasets with 100 variables due to its assumption on additive noise being
violated in discrete datasets, which is further amplified by high-dimensional
settings.

We also studied the effectiveness of \tool\ with different client sizes
($K\in\{2,4,8,16,32,64\}$) under a fixed variable size ($d=50$) in
\F~\ref{fig:rq1}. With the growth of client size, most algorithms show an
increasing trend in terms of SHD. However, \tool\ consistently has the lowest
SHD with a mild increase of processing time. In contrast, local causal graph
learning-based methods have notable difficulty in handling large client sizes
due to the low stability of local datasets and reaching a high-quality consensus
on the global causal graph.

\textbf{Answer to RQ1:}~\textit{\tool\ effectively recovers causal graphs from
federated datasets with high accuracy for varying variable sizes and client
sizes, outperforming existing methods and having negligible utility loss.}

\subsection{Resiliency}

We evaluate the performance of different algorithms in federated learning with
respect to client dropouts and heterogeneous datasets.

\begin{figure}[!htbp]
  \vspace{-10pt}
  \centering
  \begin{subfigure}[b]{0.6\columnwidth}
    \centering
    \includegraphics[width=\columnwidth]{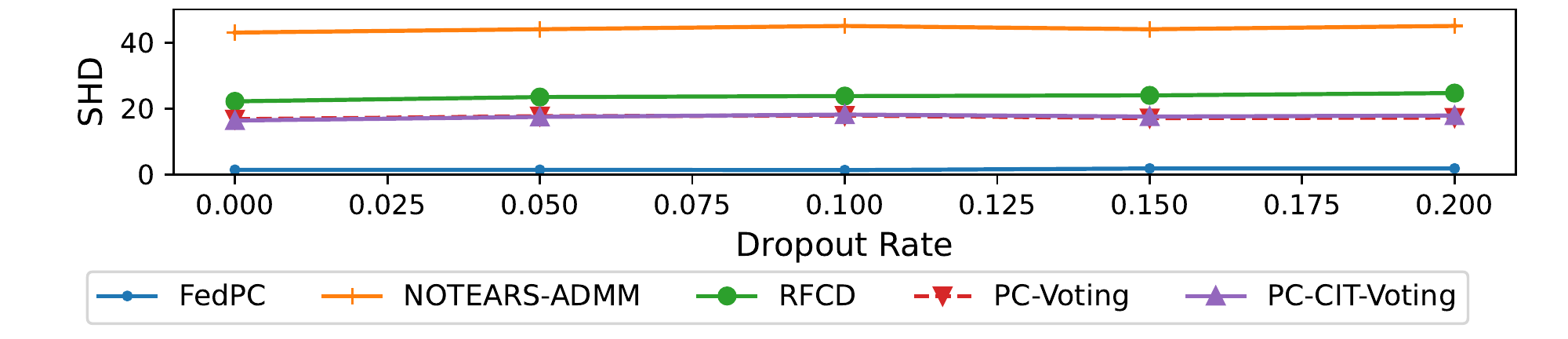}
    \caption{Resiliency to client dropouts.}
    \label{fig:dropout}
\end{subfigure}
\hfill
\begin{subfigure}[b]{0.39\columnwidth}
\vspace{10pt}
    \centering
    \includegraphics[width=\columnwidth]{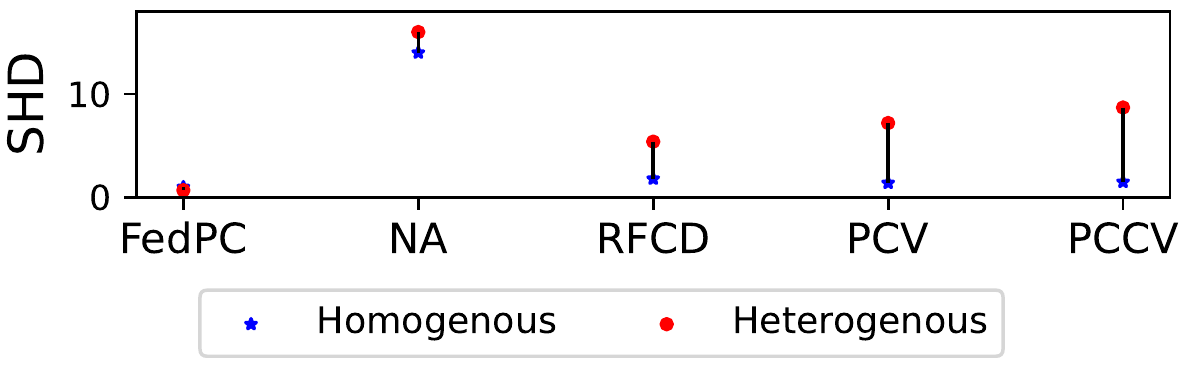}
    \caption{Resiliency to client heterogeneity.}
    \label{fig:hetero}
\end{subfigure}
\caption{Resiliency evaluation.}
\label{fig:resiliency}
\vspace{-10pt}
\end{figure}

In terms of resiliency to client dropouts, most algorithms, including our \tool,
do not explicitly consider it in their design. However, our experiments show
that the SHD of \tool\ and other algorithms does not notably downgrade even if
$20\%$ clients drop out. This emphasizes the robustness of de facto causal
structure learning algorithms to client dropouts.

Regarding resiliency to client heterogeneity, \fedpc\ performs
consistently well in both homogeneous and heterogeneous datasets ($d=20,K=4$).
Notably, \fedpc\ demonstrates negligible performance degradation 
in the presence of client heterogeneity, while other solutions,
such as NOTEARS-ADMM and RFCD, suffer notable increases in SHD (on average,
$4.7$ increase on SHD). This limitation results from their 
assumption that local datasets accurately represent the joint probability
distribution, which is invalid under heterogeneity. Actually, the local
causal graph could arbitrarily diverge from the true causal graph.

\textbf{Answer to RQ2:}~\textit{\tool\ shows resilience to both client dropouts
and client heterogeneity. Compared to other solutions, \tool\ consistently
performs well in homogeneous and heterogeneous datasets.}

\subsection{Real-world Data}

\begin{wrapfigure}{r}{0.55\textwidth}
  \vspace{-10pt}
  \centering
\includegraphics[width=0.55\columnwidth]{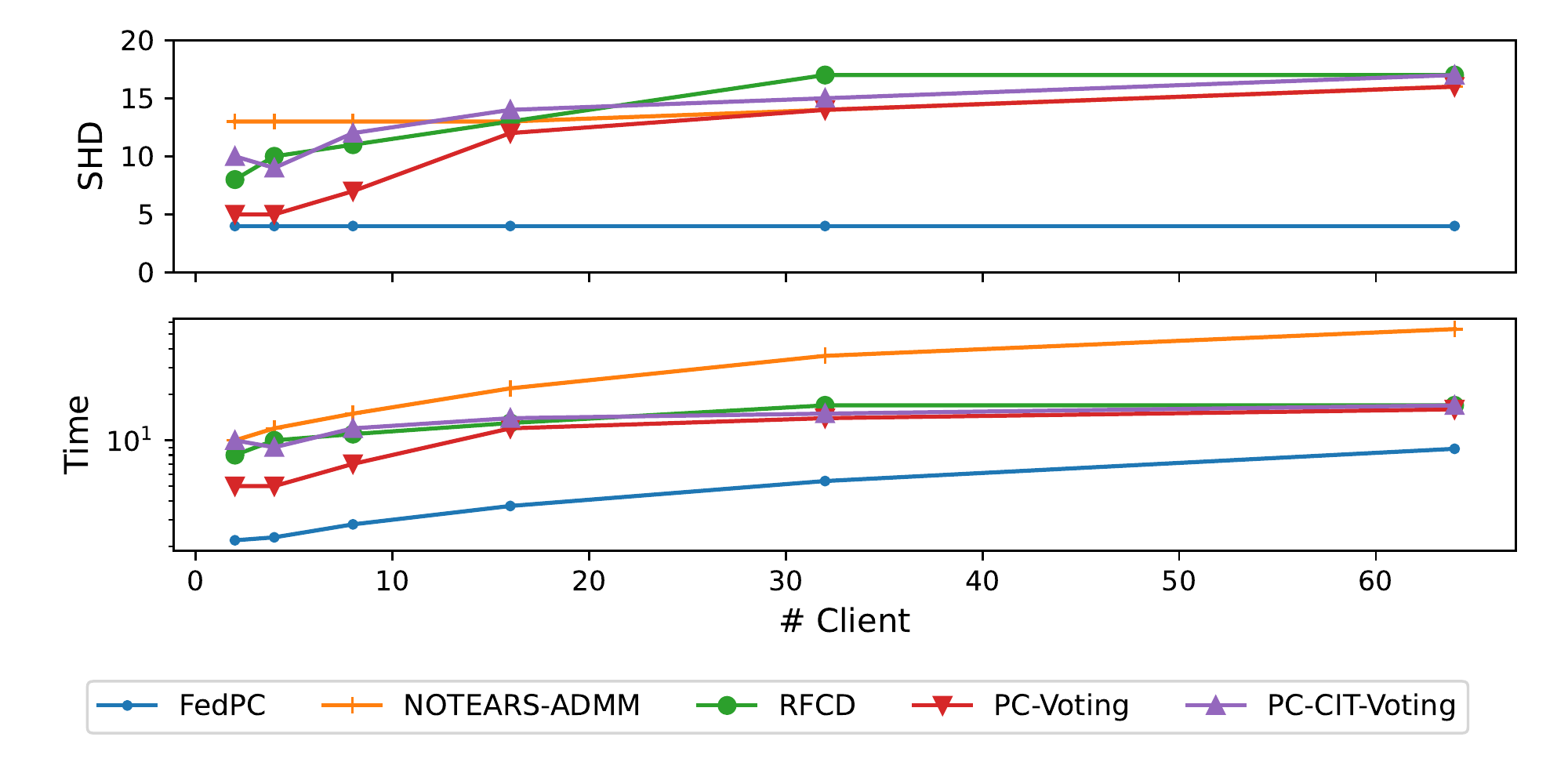}
\vspace{-5pt}
\caption{Performance on the Sachs dataset.}
\label{fig:sachs}
\vspace{-15pt}
\end{wrapfigure}

We evaluate \tool's performance on the protein expression dataset from the
real-world dataset, Scahs~\cite{sachs2005causal}, which contains $853$ samples
and $11$ variables with a ground-truth causal graph having $17$ edges. We split
the dataset into $K\in\{2,4,8,16,32,64\}$ clients and perform federated causal
structure learning. Each algorithm runs ten times for each setting and we report
the average results in \F~\ref{fig:sachs}. The results show that \tool\
demonstrates the best and highly stable performance on this dataset compared to
other algorithms. 

With 64 clients, \tool\ obtains a minimal SHD of $5.6$ while the minimal SHD of
other algorithms is $15.7$. This indicates that most edges are incorrect in
causal graphs learned by previous algorithms. We interpret that \tool\ offers a
unique advantage on learning from federated small datasets. 

\begin{figure}[!htbp]
  \centering
\includegraphics[width=\columnwidth]{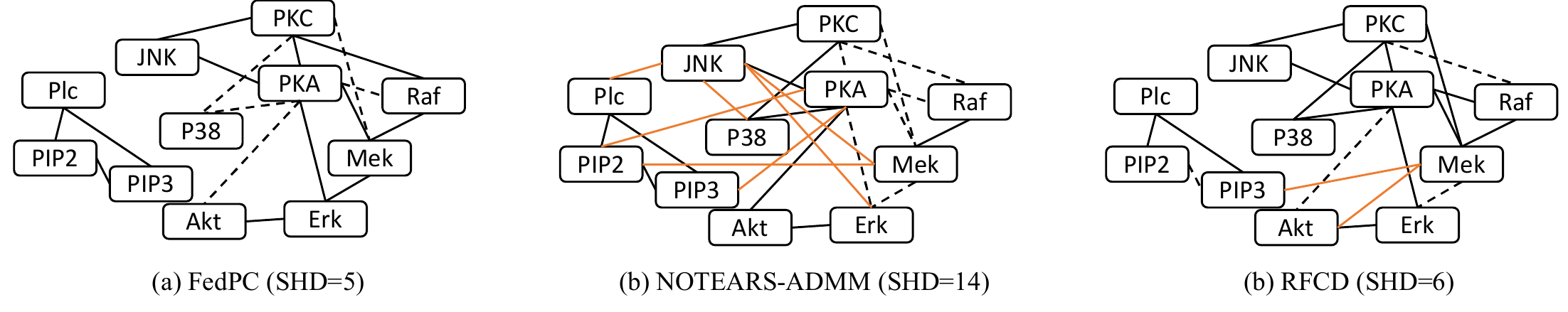}
\vspace{-10pt}
\caption{Causal graphs learned by \fedpc, NOTEARS-ADMM, and RFCD. Black solid
lines denote correct edges learned by the algorithm; orange lines denote
erroneous edges learned by the algorithm; dashed lines denote missing edges.}
\label{fig:graph}
\vspace{-15pt}
\end{figure}

We present the Markov equivalence classes of causal graphs learned by \fedpc,
NOTEARS-ADMM, and RFCD under two clients in \F~\ref{fig:graph}. In general,
\fedpc\ generates the most accurate causal graph with the lowest SHD without any
erroneous edge. In contrast, both NOTEARS-ADMM and RFCD have incorrect edges,
and NOTEARS-ADMM generates a considerable number of erroneous edges, which would
significantly undermine human comprehension of the underlying causal mechanisms
behind the data.

\textbf{Answer to RQ3:}~\textit{\tool\ outperforms other methods on the
real-world dataset, Sachs, demonstrating the best and highly stable performance
with a much lower SHD.}

\section{Related Work}

\parh{Private Causal Inference.}~Several studies have focused on privacy
protections in the causal inference process. Xu et
al.~\cite{xu2017differential}, Wang et al.~\cite{wang2020towards}, and Ma et
al.~\cite{ma2022noleaks} independently propose differentially private causal
structure learning methods. Kusner et al.~\cite{kusner2016private} present a
differentially private additive noise model for inferring pairwise cause-effect
relations, while Niu et al.~\cite{niu2022differentially} propose a
differentially private cause-effect estimation algorithm. Murakonda et
al.~\cite{murakonda2021quantifying} study the privacy risks of learning causal
graphs from data.

\parh{Federated Statistical Tests.}~The federated $\chi^2$
test~\cite{wang2021fed} is closely related to our work. It is a federated
\textit{correlation test}, whereas the $\chi^2$-test in \tool\ is designed for
\textit{conditional independence test}. Our work applies federated statistical
tests to enable practical federated causal structure learning, a crucial step in
understanding the causal relations of data and enabling causal inference.
Bogdanov et al.~\cite{bogdanov2014privacy} design an MPC-based federated
Student's t-test protocol, while Yue et al.~\cite{yue2022federated} propose a
federated hypothesis testing scheme for data generated from a linear model.
Furthermore, Gaboardi et al.~\cite{gaboardi2018local} use local differential
privacy to secure the $\chi^2$-test, and Vepakomma et
al.~\cite{vepakomma2022private} propose a differentially private independence
testing across two parties.

\parh{Federated Machine Learning.} Federated learning refers to the process of
collaboratively training a machine learning model from distributed datasets
across clients and has been studied extensively~\cite{kairouz2021advances}.
McMahan et al.~\cite{mcmahan2017communication} originally coined the term, and
since then, there have been various
proposals~\cite{yang2019federated,karimireddy2020scaffold,t2020personalized,he2020fedml,khan2020federated}
to address practical challenges, such as communication costs and non-IID data
across different clients. These proposals include update
quantization~\cite{konevcny2016federatedb,amiri2020federated}, fine-tuning
homomorphic encryption precision~\cite{zhang2020batchcrypt}, and optimizing
non-IID data~\cite{chai2020fedat,wang2020optimizing,mhaisen2020analysis}.

\section{Conclusion}

In this paper, we propose \tool, a federated constraint-based causal structure
learning framework. \tool\ is the first work that applies federated conditional
independence test protocol to enable federated causal structure learning and is
tolerant to client heterogeneity. We instantiate two algorithms with \tool,
namely \fedpc\ and \fedfci, to handle different assumptions about data. Through
extensive experiments, we find \tool\ manifests competitive performance on both
synthetic data and real-world data.

\section*{Acknowledgement}

We thank the anonymous reviewers for their insightful comments. We also thank Qi
Pang for valuable discussions. This research is supported in part by the HKUST
30 for 30 research initiative scheme under the the contract Z1283 and the
Academic Hardware Grant from NVIDIA.

%
%
%
\bibliographystyle{splncs04}
\bibliography{main,ref}

\clearpage

\section*{Supplementary Material}

\subsection{Workflow}

\begin{figure}[!htbp]
  \vspace{-10pt}
  \centering
\includegraphics[width=0.6\columnwidth]{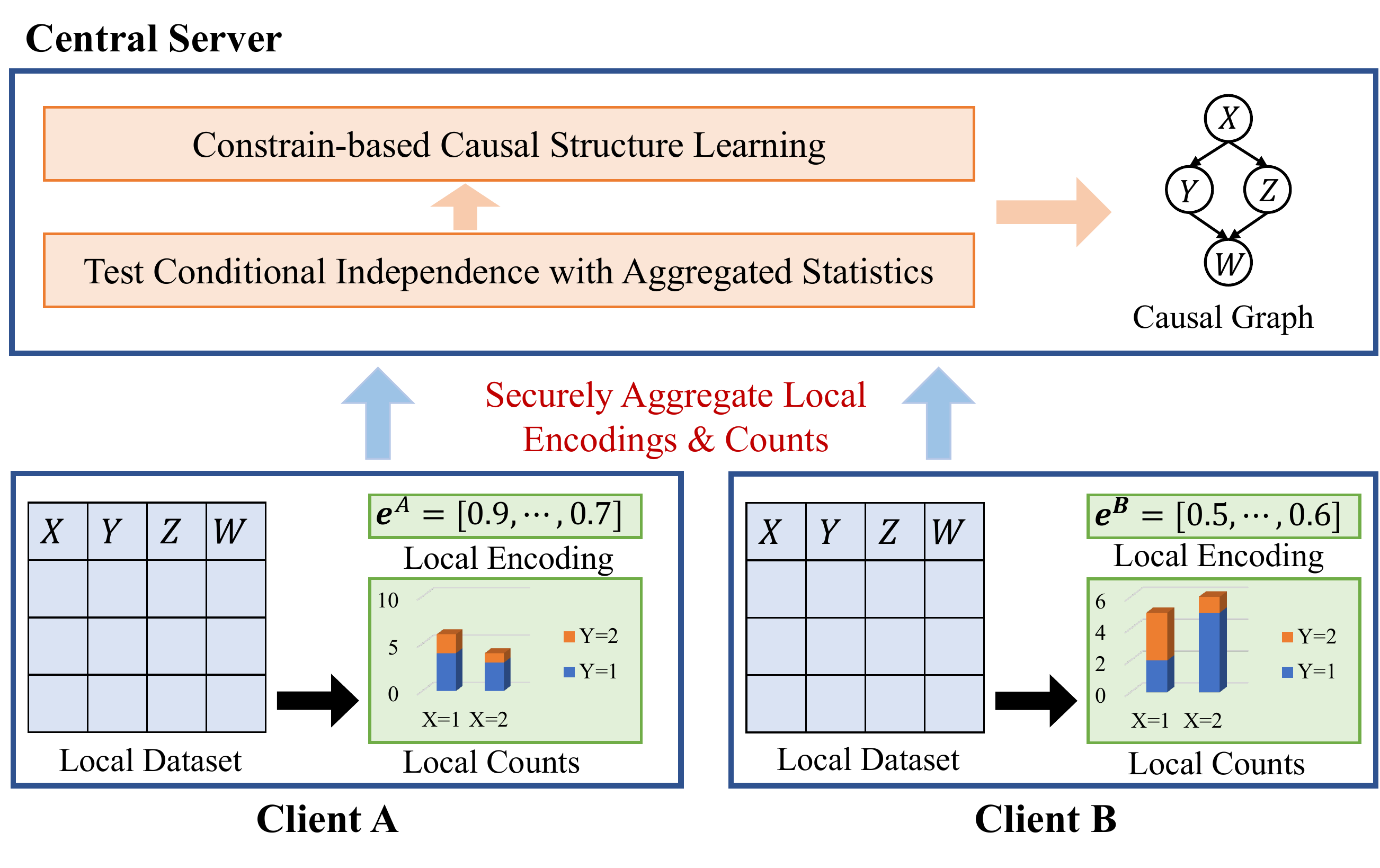}
\vspace{-10pt}
\caption{\tool\ workflow.}
\label{fig:workflow}
\vspace{-15pt}
\end{figure}

\subsection{Graph Separation}

\begin{definition}[d-separation]
    Two nodes $X,Y$ are d-separated by a set of nodes $\bm{Z}$ in a causal graph
    $G$ if and only if $X,Y$ are not d-connected by $\bm{Z}$ in $G$. Two nodes
    $X,Y$ are d-connected by $\bm{Z}$ in $G$ if and only if an undirected path
    $U$ connects them in such a way that for each collider $W$ on $U$, either
    $W$ or a descendant of $W$ is in $\bm{Z}$, and no non-collider on $U$ is in
    $\bm{Z}$. If $\to W \leftarrow$ exist in path $U$ ($\to$ and $\leftarrow$
    are directed edges), $W$ is a collider.
\end{definition}

\begin{definition}[m-separation]
    Two nodes $X,Y$ are m-separated by a set of nodes $\bm{Z}$ in a causal graph
    $G$ if and only if there is no active path between $X,Y$ relative to
    $\bm{Z}$ in $G$. A path $U$ between $X$ and $Y$ is active relative to a
    (possible empty) set of nodes $\bm{Z}$ if (i) every non-collider on $U$ is
    not a member of $\bm{Z}$ and (ii) every collider on $U$ has a descendant in
    $\bm{Z}$.
\end{definition}

\clearpage

\subsection{PC Algorithm}

\begin{algorithm}[H]
\small
\caption{PC Algorithm}
\label{alg:csl}
\SetKwFunction{Skl}{Skeleton-Learning}
\SetKwFunction{PC}{PC}
\SetKwFunction{FCI}{FCI}
\SetKwProg{Fn}{Function}{:}{}
\Fn{\Skl{$X_1,\cdots,X_d$}}{
    let $S_0$ be a complete graph over $X_1,\cdots,X_d$\;
    let $S$ be a copy of $S_0$\;
    \ForEach{edge $(X,Y)$ in $S_0$}{
        $V\leftarrow \{X_1,\cdots,X_d\}\setminus \{X,Y\}$\;
        \uIf{$\exists \bm{Z}\subseteq V.  X\ci Y\mid \bm{Z}$}{
            remove $(X,Y)$ in $S$;
        }
    }
    \KwRet{$S$}
}

\Fn{\PC{$X_1,\cdots,X_d$}}{
    \algcomment{// skeleton learning\\}
    $G\leftarrow \Skl(X_1,\cdots,X_d)$\;
    \algcomment{// orientation\\}
    \ForEach{unshielded triple $(X-Z-Y)$ in $G$}{
        \uIf{$\forall X\ci Y\mid G, Z\notin G$}{
            orient $(X-Z-Y)$ as $(X\to Z\leftarrow Y)$;
        }
    }
    \Repeat{no more edgs can be oriented}{
        \lForEach{$(X\to Z-Y)$}{orient as $(X\to Z\leftarrow Y)$}        
        \lForEach{$(X\to Z\to Y)$ and $(X-Y)$}{orient as $(X\to Y)$}        
        \lForEach{$(X-Y)$, $(X-Z)$, $(X-W)$, $(Z\to Y)$, $(W\to Y)$ and $Z,W$ is nonadjacent}{orient as $(X\to Y)$}        
    }
    \KwRet{$G$}
}
\end{algorithm}

We present the workflow of the PC algorithm~\cite{spirtes2000causation} in
\A~\ref{alg:csl}.~\footnote{We use a simplified skeleton learning algorithm in
\A~\ref{alg:csl} for the ease of presentation.} In the first step (lines 1--9;
line 12), edge adjacency is confirmed if there is no conditional independence
between two variables (line 6). In the second step (lines 14--22), a set of
orientation rules are applied based on conditional independence and graphical
criteria.

\subsection{Dataset and Hyperparameters}

For the synthetic datasets, we use the Erdős–Rényi (ER) random graph
model~\cite{ergraph} to synthesize DAGs with $d\in\{10, 20, 50, 100\}$ nodes. We
sample the graph parameters from a Dirichlet-multinomial distribution. To
generate causally insufficient datasets, we randomly mask some variables in the
DAG and generate its corresponding MAG. We use forward sampling to obtain 10,000
samples per dataset, which we split into $K\in\{2,4,8,16,32,64\}$ partitions as
local datasets. To generate client heterogeneous datasets, we add a surrogate
variable to the DAG and split the dataset for $K$ clients according to its value
(\D~\ref{def:hetro}).

We set the hyperparameters of our evaluations as follows. The encoding size $l$
in \Thm~\ref{thm:encoding} is $50$. The $\alpha$ of \A~\ref{alg:cit} is $0.05$.
NOTEARS-ADMM is concretized with Multi-Layer Perceptron (MLP) to handle the
non-linearity in data. We use the default parameters in NOTEARS-ADMM and perform
a grid search on the threshold $\tau$ with best performance. RFCD is concretized
with the GES~\cite{spirtes2000causation} algorithm with the BDeu score
function~\cite{buntine2014theory} that is particularly designed for discrete
data. We use standard $\chi^2$-test with $\alpha=0.05$ in PC, PC-Voting,
PC-CIT-Voting, FCI, FCI-Voting, and FCI-CIT-Voting.

\subsection{Evaluation on Downstream Application}

Following~\cite{ma2022noleaks}, we also launch \tool\ in a downstream task of
causal structure learning, namely, causal feature selection (CFS). Regression or
classification models often suffer out-of-distribution (OOD) issues, which
undermines their accuracy on test data. Here, OOD indicate that the distribution
of test data is different from the distribution of training data. From the
perspective of causality, OOD issues can be interpreted by causal mechanism
shifts in the underlying causal graph. Such shifts, however, is prevalent in
practice in the evolving environments. CSF is motivated by the observation that
direct causal relations are often more reliable against OOD than indirect causal
relations. Hence, CFS aims to pick a subset of variables with strong causal
relations to improve the robustness of models on OOD data, instead of using full
features. 

\begin{table}[!htbp]
  \centering
  \caption{Performance in the LUCAS benchmark. Best test MSE is
  \underline{highlighted}. ``\#Features'' denotes the number of remaining
  features after feature selection.} \resizebox{0.9\linewidth}{!}{
  \begin{tabular}{l|l|c|c|c}
    \hline
    \multicolumn{5}{c}{\#Client=2}\\
    \hline
    \textbf{Dataset} & \textbf{Method} & \textbf{\#Features}& \textbf{Training MSE}& \textbf{Test MSE}\\
    \thickhline
    \multirow{6}{*}{LUCAS 1} & None & 11 & 0.423 & 0.642 \\\cline{2-5}
    & \fedpc& 4 & 0.445 & \underline{0.608} \\\cline{2-5}
    & NOTEARS-ADMM & 3 & 0.457 & 0.631 \\\cline{2-5}
    & RFCD & 4 & 0.435 & 0.656 \\\cline{2-5}
    & PC-Voting & 4 & 0.445 & \underline{0.608} \\\cline{2-5}
    & PC-CIT-Voting & 3 & 0.457 & 0.631 \\\thickhline
    \multirow{6}{*}{LUCAS 2} & None & 11 & 0.423 & 0.631 \\\cline{2-5}
    & \fedpc& 4 & 0.445 & \underline{0.574} \\\cline{2-5}
    & NOTEARS-ADMM & 3 & 0.457 & 0.589 \\\cline{2-5}
    & RFCD & 4 & 0.435 & 0.646 \\\cline{2-5}
    & PC-Voting & 4 & 0.445 & \underline{0.574} \\\cline{2-5}
    & PC-CIT-Voting & 3 & 0.457 & 0.589 \\\hline

    \multicolumn{5}{c}{\#Client=4}\\
    \hline
    \multirow{6}{*}{LUCAS 1} & None & 11 & 0.423 & 0.642 \\\cline{2-5}
    & \fedpc& 4 & 0.445 & \underline{0.608}\\\cline{2-5}
    & NOTEARS-ADMM & 4 & 0.435 & 0.656 \\\cline{2-5}
    & RFCD & 3 & 0.457 & 0.631 \\\cline{2-5}
    & PC-Voting & 4 & 0.445 & \underline{0.608 }\\\cline{2-5}
    & PC-CIT-Voting & 3 & 0.457 & 0.631 \\\thickhline
    \multirow{6}{*}{LUCAS 2} & None & 11 & 0.423 & 0.631 \\\cline{2-5}
    & \fedpc& 4 & 0.445 & \underline{0.574} \\\cline{2-5}
    & NOTEARS-ADMM & 4 & 0.435 & 0.646 \\\cline{2-5}
    & RFCD & 3& 0.457 & 0.589 \\\cline{2-5}
    & PC-Voting & 4 & 0.445 & \underline{0.574} \\\cline{2-5}
    & PC-CIT-Voting & 3 & 0.457 & 0.589 \\\hline
    \multicolumn{5}{c}{\#Client=8}\\
    \hline
    \multirow{6}{*}{LUCAS 1} & None & 11 & 0.423 & 0.642 \\\cline{2-5}
    & \fedpc& 4 & 0.445 & \underline{0.608} \\\cline{2-5}
    & NOTEARS-ADMM & 6 & 0.443 & 0.628 \\\cline{2-5}
    & RFCD & 2 & 0.480 & 0.616 \\\cline{2-5}
    & PC-Voting & 3 & 0.457 & 0.631 \\\cline{2-5}
    & PC-CIT-Voting & 3 & 0.457 & 0.631 \\\thickhline
    \multirow{6}{*}{LUCAS 2} & None & 11 & 0.423 & 0.631 \\\cline{2-5}
    & \fedpc& 4 & 0.445 & \underline{0.574} \\\cline{2-5}
    & NOTEARS-ADMM & 6 & 0.443 & 0.635 \\\cline{2-5}
    & RFCD & 2 & 0.480 & 0.590 \\\cline{2-5}
    & PC-Voting & 3 & 0.457 & 0.589 \\\cline{2-5}
    & PC-CIT-Voting & 3 & 0.457 & 0.589 \\\hline
  \end{tabular}
   }
\label{tab:lucas}
\end{table}

We use the LUCAS dataset~\cite{cawley2008causal} that is particularly designed
for assessing causal feature selection algorithms. This dataset contains a
training set and two OOD test datasets---LUCAS1 and LUCAS2. LUCAS2 suffers from
a higher degree of domain shifts (i.e., more ``OOD'') than that of LUCAS1. We
use different federated causal structure learning algorithms to select the
subset of features, perform model training on the selected features, and measure
the regression errors in the form of Mean Square Error (MSE) on each test
dataset (lower is better). The results are shown in \T~\ref{tab:lucas}. We
observe that \fedpc\ consistently finds the best features regardless of the
client sizes. However, PC-CIT-Voting, while showing comparable performance with
\#Client=4/8, fails to identify the subset of features with strong causal
relations when \#Client reaches to eight. Other methods either over-aggressively
rule out useful features or use too many features that undermine the performance
on the OOD data.

\end{document}